\documentclass{article}

\usepackage{authblk}
\usepackage{times}
\usepackage{epsfig}
\usepackage{graphicx}
\usepackage{amsmath}
\usepackage{amssymb}
\usepackage{enumerate}
\usepackage{paralist}
\usepackage{cite}
\usepackage[font=small,labelfont=bf,tableposition=top]{caption}

\usepackage{tikz}

\usepackage{accents}
\newlength{\dhatheight}

\usepackage{array}
\newcolumntype{P}[1]{>{\centering\arraybackslash}p{#1}}

\usepackage{tabularx}

\usepackage[pagebackref=true,breaklinks=true,letterpaper=true,colorlinks,bookmarks=false]{hyperref}

\begin{document}

\title{Adapting Convolutional Neural Networks for Geographical Domain Shift}

\author[1]{Pavel Ostyakov}
\author[1,2,3]{Sergey I. Nikolenko}

\affil[1]{Samsung AI Center, Moscow, Russia}
\affil[2]{Steklov Institute of Mathematics at St. Petersburg, Russia}
\affil[3]{Neuromation OU, Tallinn, Estonia}

\date{\vspace{-5ex}}

\maketitle


\begin{abstract}
We present the winning solution for the Inclusive Images Competition organized as part of the Conference on Neural Information Processing Systems (NeurIPS 2018) Competition Track. The competition was organized to study ways to cope with domain shift in image processing, specifically geographical shift: the training and two test sets in the competition had different geographical distributions. Our solution has proven to be relatively straightforward and simple: it is an ensemble of several CNNs where only the last layer is fine-tuned with the help of a small labeled set of tuning labels made available by the organizers. We believe that while domain shift remains a formidable problem, our approach opens up new possibilities for alleviating this problem in practice, where small labeled datasets from the target domain are usually either available or can be obtained and labeled cheaply.
\end{abstract}

\section{Introduction}

\emph{Domain shift} is a problem often arising in machine learning, when a model is trained on a dataset that might be sufficiently large and diverse, but later the model is supposed to be applied to datasets with a different data distribution. One important example of this problem is the \emph{geographical} domain shift in image processing, when, e.g., the same semantic category of objects can look quite different on photos taken in different geographical locations (see Fig.~\ref{fig:examples}). Domain shift also often results from \emph{dataset bias}: e.g., a dataset of human faces heavily shifted towards Caucasian faces would suffer from this problem when applied in, e.g., Asia.

Modern techniques in domain adaptation (see references in Section~\ref{sec:related}) usually operate in conditions where the target domain is completely different from the source domain in some aspects; e.g., the source domain are synthetic images generated artificially and the target domain includes the corresponding real images. Geographical domain shift is a more subtle problem: in an image classification problem with geographical shift, some classes will not change at all from the source to target domain, while others might change radically.

In this work, we present the winning solution for the Inclusive Images Competition~\cite{challenge} organized as part of the Conference on Neural Information Processing Systems (NeurIPS 2018) Competition Track. Based on the work~\cite{46553}, the challenge was intended to develop solutions for the geodiversity problem, with a training set skewed towards North America and Western Europe and test sets drawn from completely different geographic distributions that were not revealed to the competitors.

One interesting property of our solution is that it is relatively straightforward and simple. We did not use any state of the art models for domain adaptation, and our final solution is an ensemble of several CNNs where only the last layer is fine-tuned with the help of a small labeled set of tuning labels (Stage~1 set) that was made available by the organizers. It turned out that this set had a geographical distribution similar enough to the hidden Stage~2 evaluation set, and the very small set of tuning labels (only $1000$ examples) proved to suffice, with proper techniques such as data augmentation and ensembling, to adapt the base models to a new domain.

The paper is organized as follows. In Section~\ref{sec:related}, we survey some related work on domain shift and domain adaptation. Section~\ref{sec:problem} introduces the problem of the Inclusive Images Challenge and describes the dataset and evaluation metrics. Section~\ref{sec:methods} presents our solution in detail, Section~\ref{sec:experiments} shows experimental results for both single-model solutions and ensembles, and Section~\ref{sec:concl} concludes the paper.

\section{Related Work}\label{sec:related}

Over the last decade, convolutional neural networks have defined state of the art in the image classification task. While not trying to provide a comprehensive survey, we note the works that introduced image classification architectures that we use in this work: deep residual networks~\cite{He2016DeepRL}, densely connected convolutional networks~\cite{8099726}, architectures produced by neural architecture search~\cite{Zoph2018LearningTA} and progressive neural architecture search~\cite{Liu2018ProgressiveNA}, and squeeze-and-excitation networks~\cite{Hu2018SqueezeandExcitationN}.

Generally speaking, \emph{domain adaptation}, i.e., adaptation of machine learning models so that they would work well on a target dataset different from the one they trained on (source dataset) has been considered in maching learning for a long time~\cite{DBLP:journals/corr/ZhangLO17,DBLP:journals/corr/abs-1802-03601,7078994,Bridle:1990:RSN:118850.118882}, including theoretical results that connect domain adaptation with the theory of $\mathcal{H}\Delta\mathcal{H}$-divergence~\cite{Ben-David:2010:TLD:1745449.1745461,NIPS2006_2972,NIPS2007_3212}. Solutions for image processing problems began to appear in the early 2010s, based either on directly training cross-domain transformations~\cite{saenko2010adapting,DBLP:journals/corr/DonahueJVHZTD13,DBLP:journals/corr/LuZCWXSH17} or on \emph{adversarial} domain adaptation, where feature extractors are trained together with the cross-domain transformation in an adversarial scheme. Many recent works use GAN-based architectures for this kind of domain adaptation~\cite{tzeng2015simultaneous,tzeng2017adversarial,ganin2015unsupervised,DirtT,liu2016coupled,ghifary2016deep}. A different adversarial approach would be to apply direct style transfer between source and target domains (\emph{domain transfer})~\cite{isola2017image,yoo2016pixel,Applerefiner_2017_CVPR,bousmalis2017unsupervised,zhu2017unpaired,Hoffman_cycada2017}. Deep transfer learning is also a very promising technique which is often used in image classification, especially for small datasets~\cite{oquab2014learning}.

As for domain shift specifically in the geographical context, in~\cite{shankar2017no} the authors analyzed geodiversity in two large-scale image datasets: ImageNet~\cite{deng2009imagenet} and OpenImages~\cite{OpenImages}. They divided images by their respective geolocations and trained image classifiers. As expected, an extreme difference was found in terms of performance on train and test sets. The work~\cite{shankar2017no} concludes that a dataset should contain images from different geographical locations in order to be applicable for real-world applications. The work~\cite{zou2018ai} also raises the problem of geodiversity lacking in standard datasets.

Diversity in existing datasets have also been explored in previous works. In \cite{vodrahalli2018all}, the authors show that the size of a dataset is not as important as the diversity of samples within it. They propose a new method to choose a subset of examples that would be sufficient for training and getting virtually identical results in terms of validation accuracy. The works~\cite{torralba2011unbiased,DBLP:journals/corr/TommasiPCT15} have uncovered and studied biases in classical image processing datasets.

\section{Problem Statement}\label{sec:problem}

\newcommand{\mypic}[2]{\includegraphics[width=#1\linewidth]{images/examples/#2.eps}}

\begin{figure}[t!]
    \setlength{\tabcolsep}{1pt}
    \centering\footnotesize
    \begin{tabular}{P{.35\linewidth}P{.3\linewidth}P{.35\linewidth}}
    \mypic{.85}{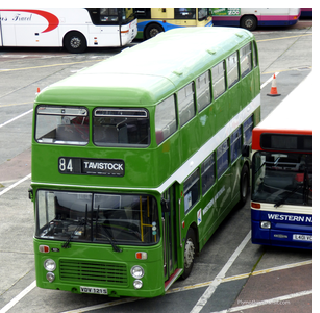} &
    \mypic{.95}{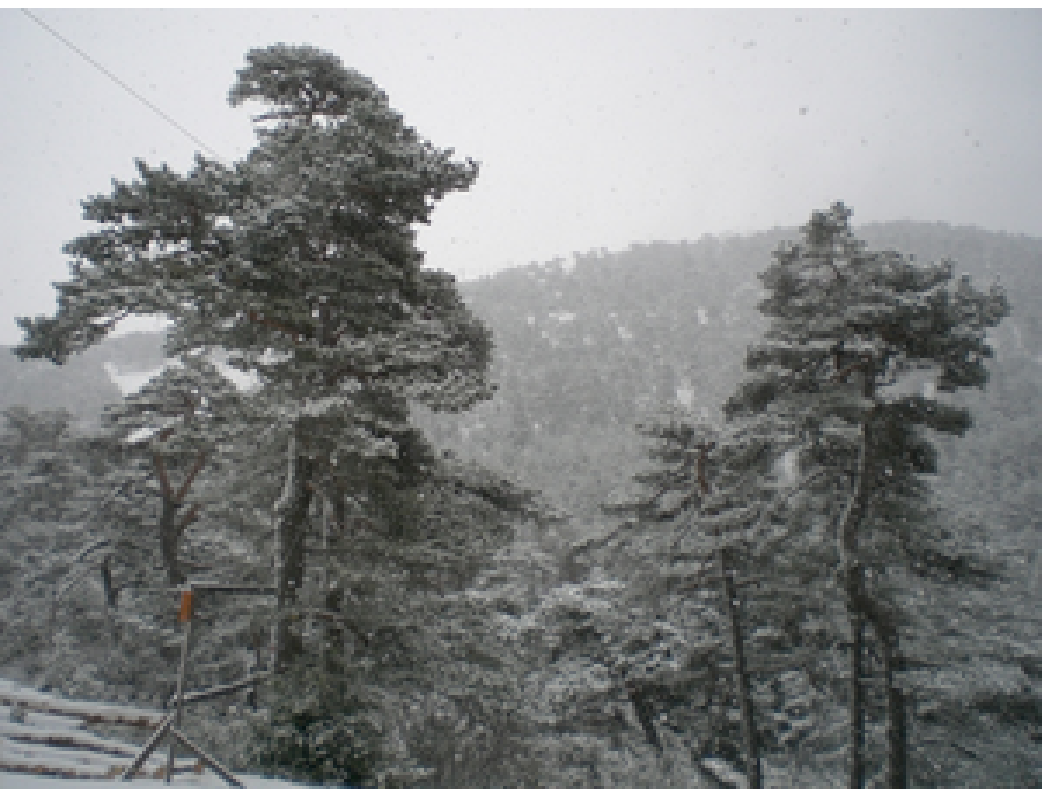} &
    \mypic{.95}{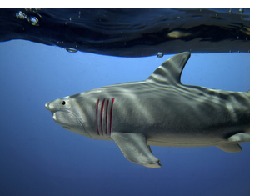} \\
    (a) Bus, Double-decker bus, Public transport, Mode of transport, Transport, Vehicle &
    (b) Snow, Tree &
    (c) Fish \\
    \mypic{.8}{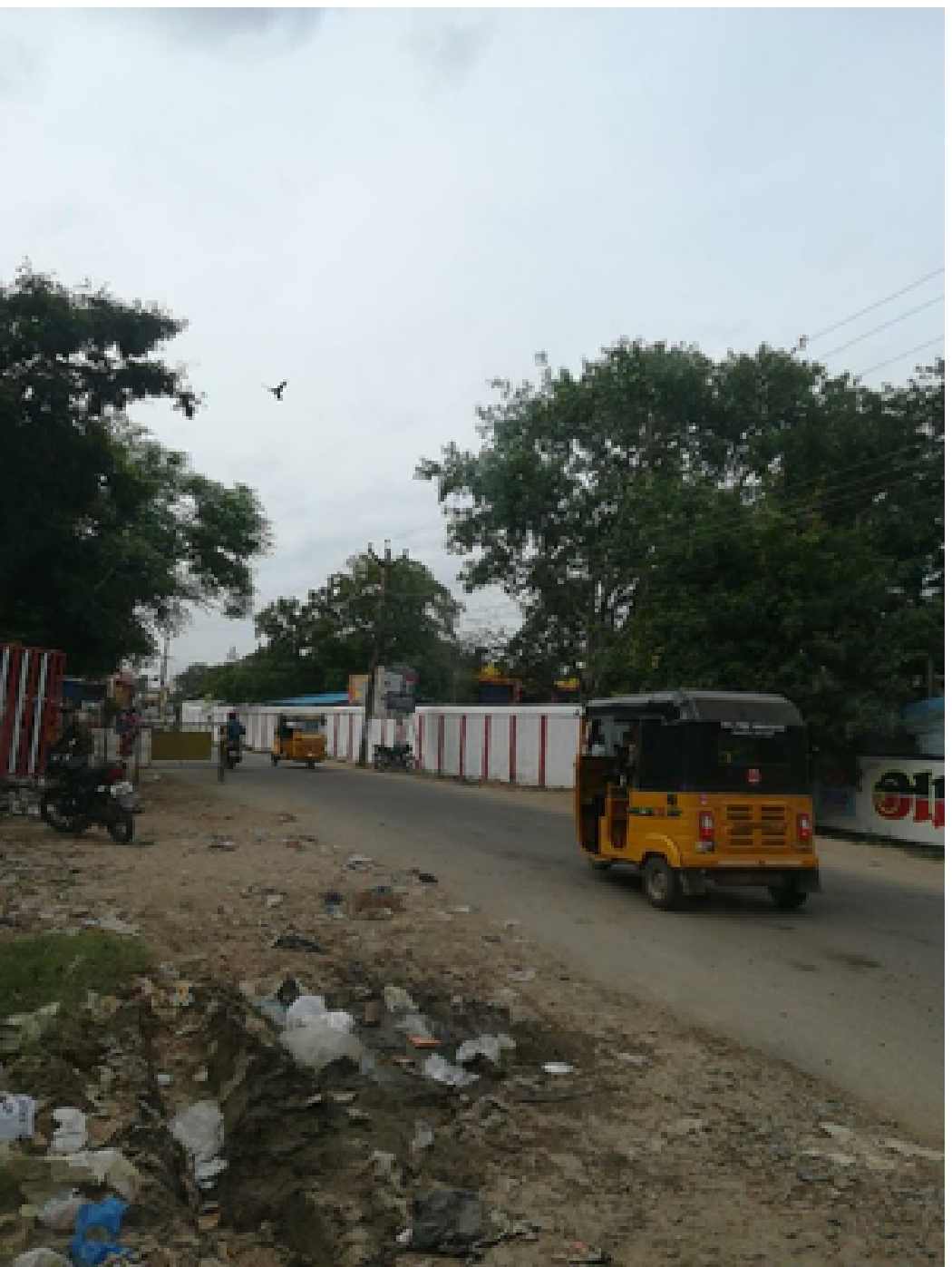} &
    \mypic{.9}{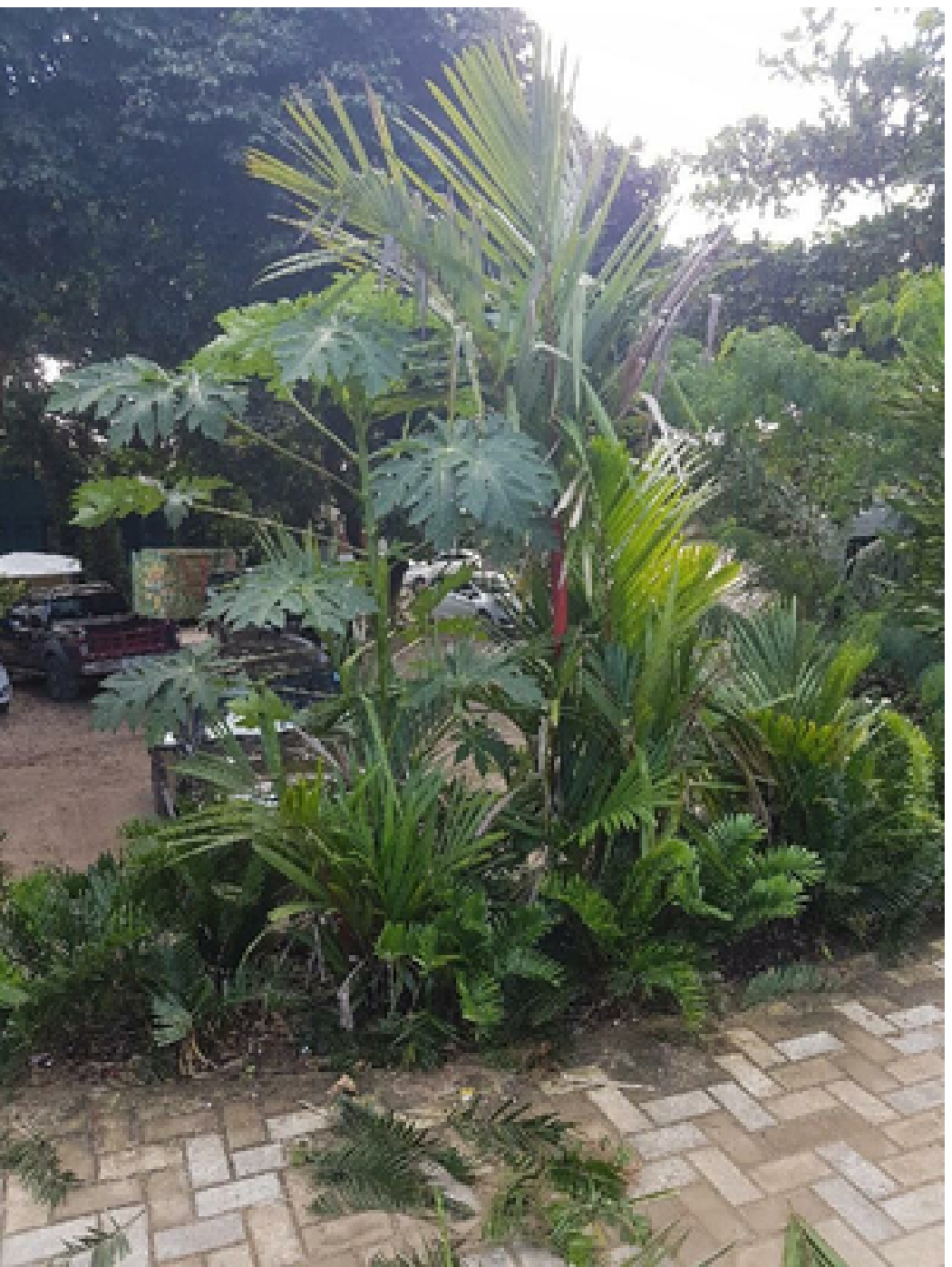} &
    \mypic{.6}{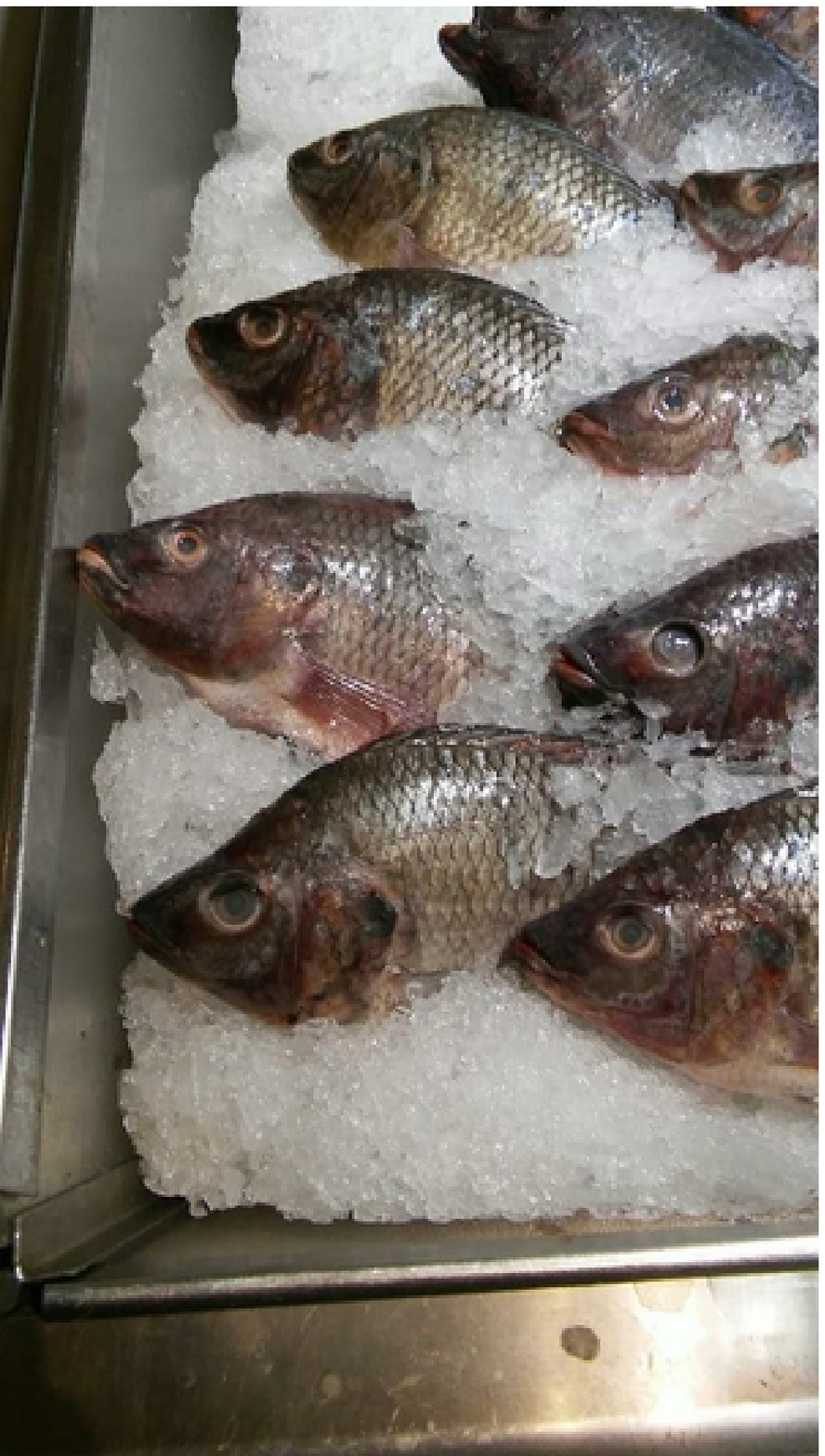} \\
    (d) Tree, Person, Transport, Public transport &
    (e) Papaya, Tree &
    (f) Fish, Meat \\
    \mypic{.6}{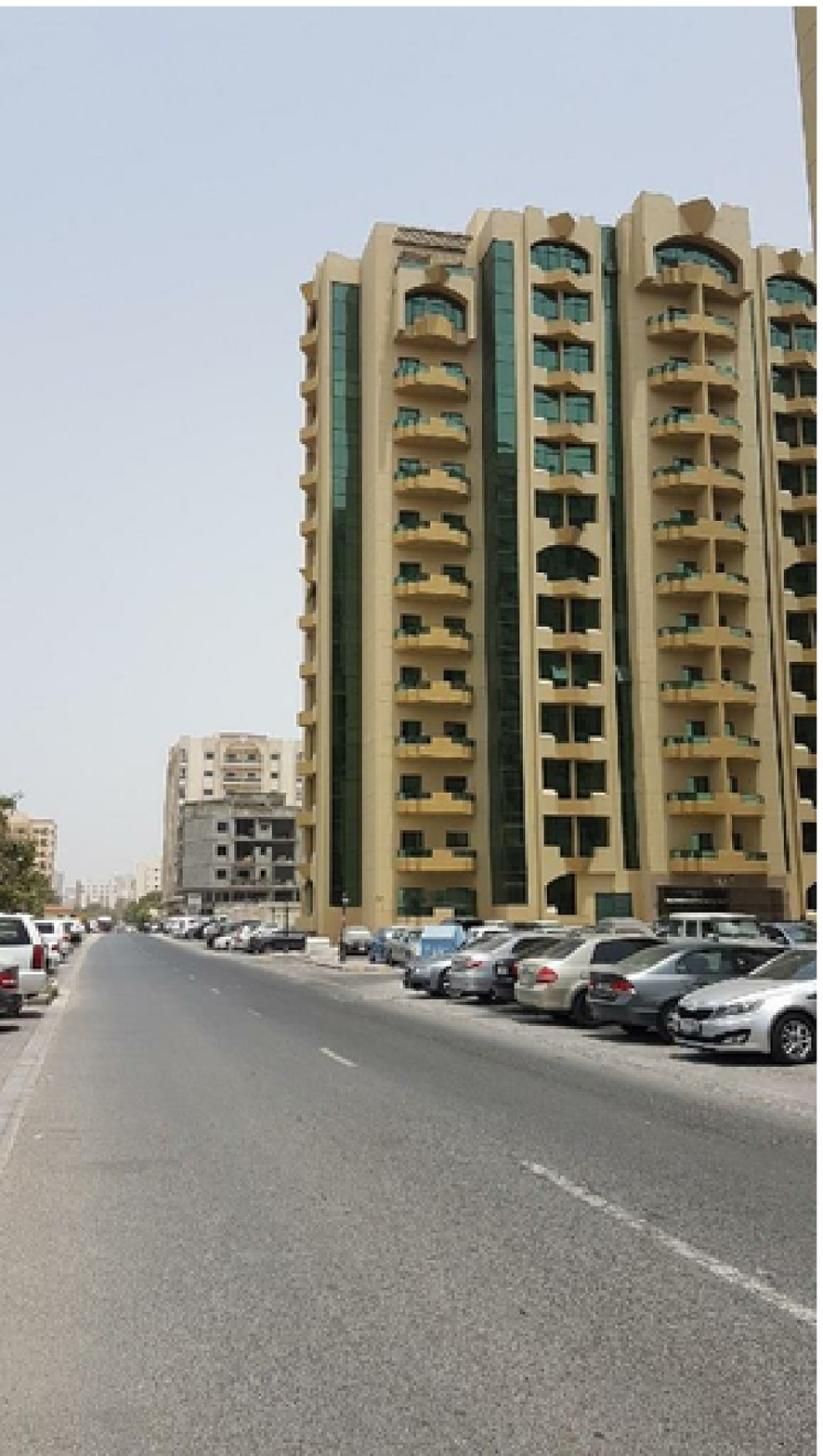} &
    \mypic{.7}{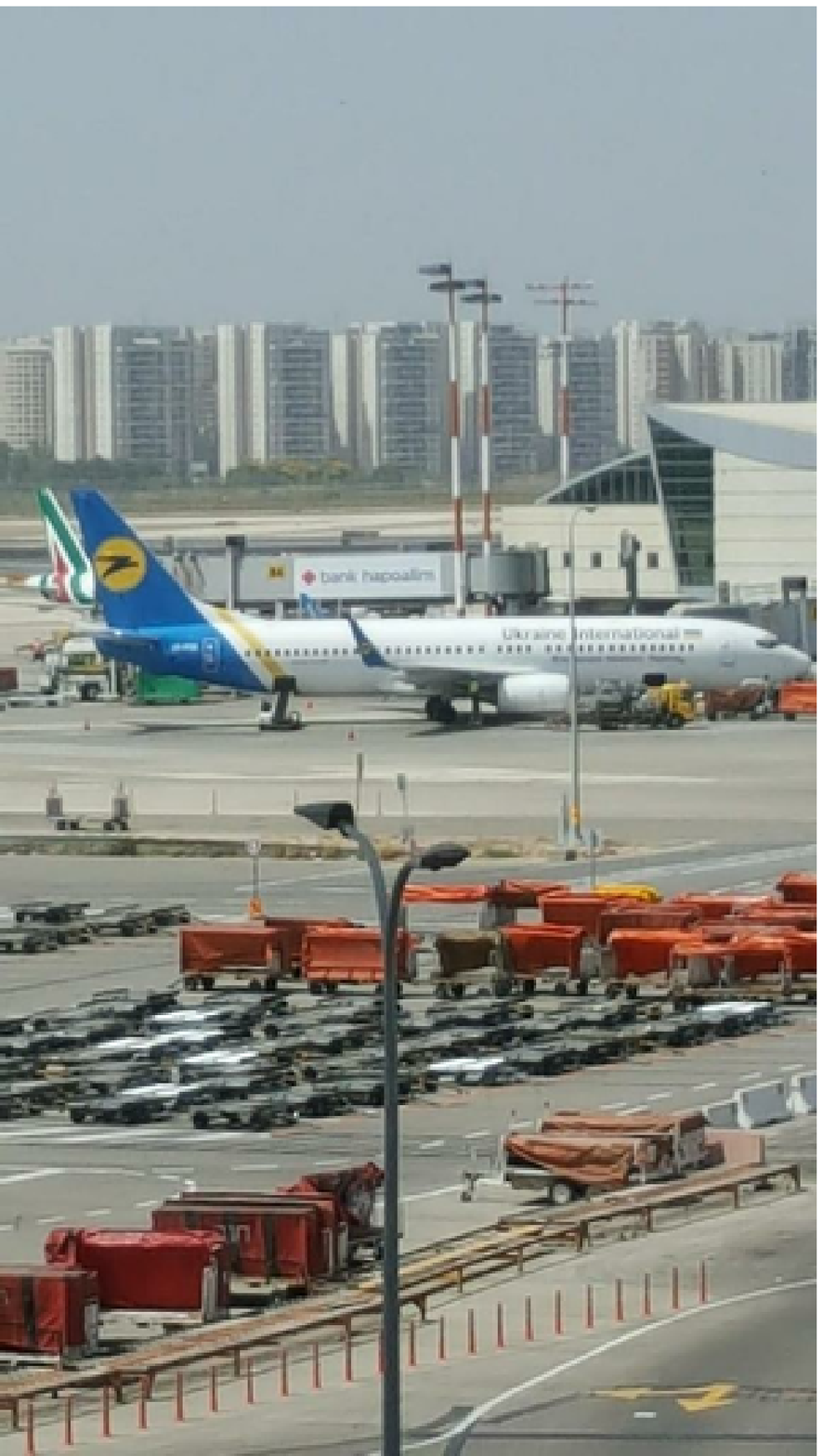} &
    \mypic{.8}{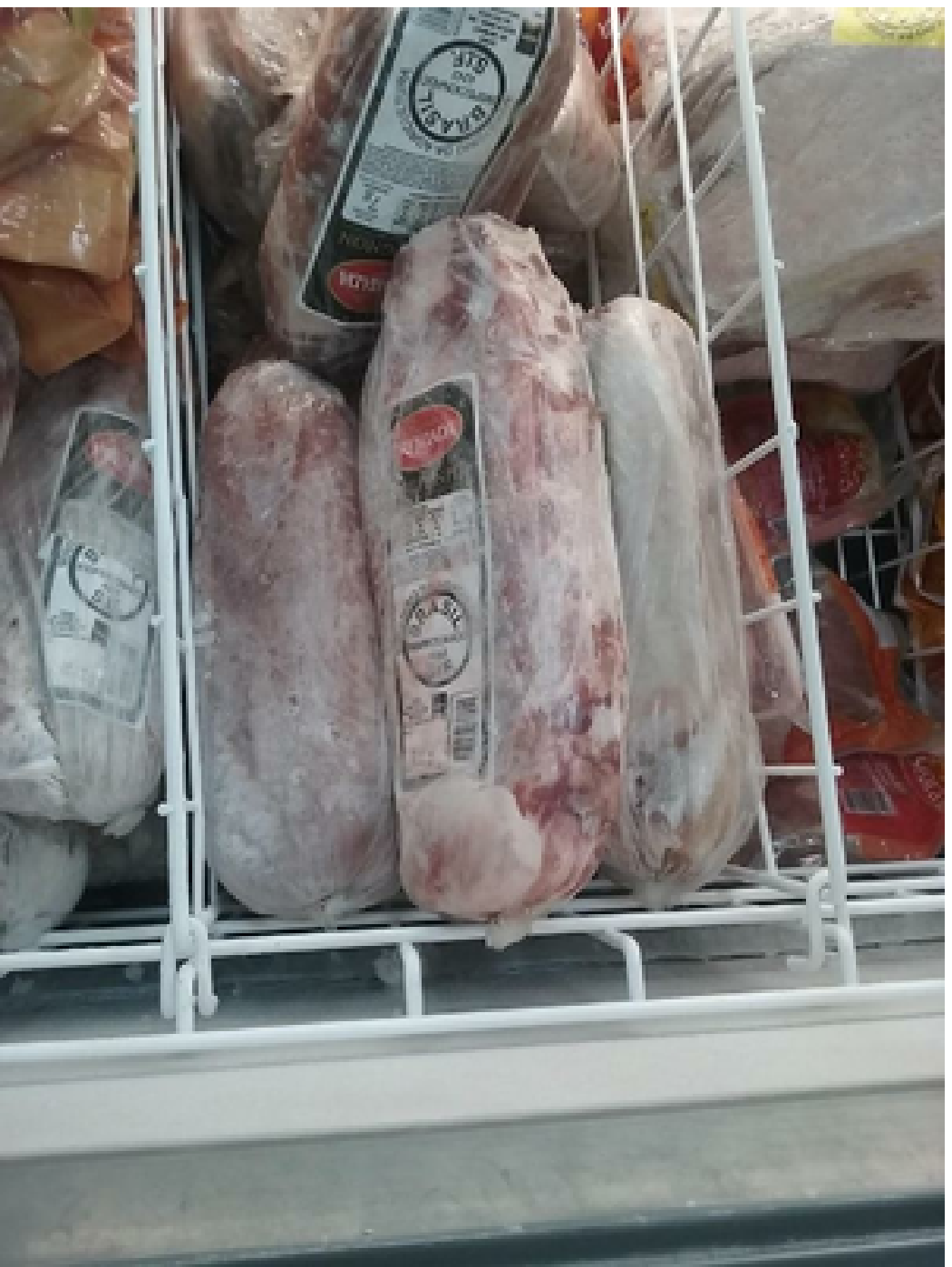} \\
    (g) Tower block, Neighbourhood, Tower, Apartment &
    (h) Aircraft, Airport, Boeing 737 next generation &
    (i) Frozen food, Meat, Pork
    \end{tabular}
    \caption{Sample images from the Inclusive Images datasets together with the corresponding labels. Top row, (a)-(c): sample images from the training set; middle and bottom rows, (d)-(f): sample images with similar labels from the Stage 1 test dataset.}\label{fig:examples} \vspace{-.1cm}
\end{figure}

\begin{figure}\centering
    \begin{tabular}{p{.32\textwidth}p{.32\textwidth}p{.32\textwidth}}
    \multicolumn{3}{c}{\includegraphics[width=\textwidth]{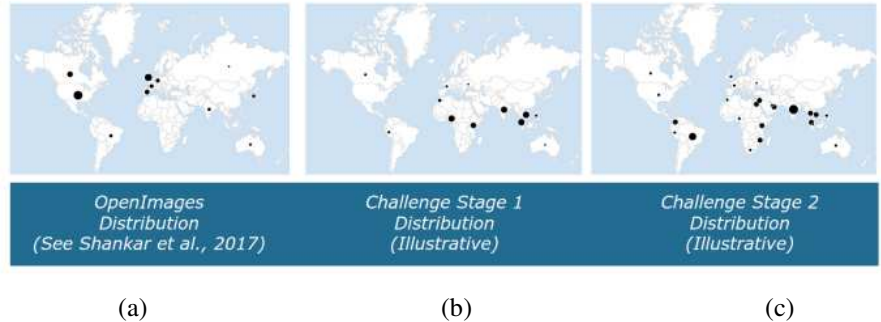}} \\
    \centering (a) & \centering (b) & \centering (c)
    \end{tabular}
   \caption{Geographical distributions of the three datasets used in the competition: (a) training set (OpenImages); (b) Challenge Stage 1 (public validation set and tuning labels); (c) Challenge Stage 2 (hidden test set); picture taken from~\cite{challenge}.}\label{fig:distribution}
\end{figure}

The Inclusive Images Challenge had two datasets released for the participants:
\begin{itemize}
    \item a training set of $1.7$ million images with ground truth labels from one geographical distribution (training set distribution; see Fig.~\ref{fig:distribution}a); the training set was taken from the \emph{OpenImages} dataset~\cite{OpenImages} for classification;
    \item a public test set of $33$ thousand images used to score the public leaderboard (Challenge Stage 1) from a different geographical distribution (Stage 1 distribution; see Fig.~\ref{fig:distribution}b); the Stage~1 test set also contained ground truth labels for $1000$ images, called \emph{tuning labels}.
\end{itemize} 
The final scores for the challenge were scored on a hidden test set that was never released for training or tuning; it contained a third geographical distribution, called Challenge Stage 2 distribution; see Fig.~\ref{fig:distribution}c. As we can see on Figure~\ref{fig:distribution}, there is a big difference between the training set geographic distribution and both Stage~1 and Stage~2 test sets. However, Stage~1 and Stage~2 distributions are very similar; this will become an important point for our solution later.

Formally, the problem is a multilabel classification problem with $7178$ classes (tags); a single photo can have several tags assigned to it. Figure~\ref{fig:examples} shows a few examples of the images from the challenge datasets. By inspecting the datasets, we have found the following properties that have proven to be important for our solution:
\begin{itemize}
    \item despite the main point made in the challenge description, the actual pictures that represent different classes do not change that much with the geographical distribution; a wide class such as ``Person'' can become even wider in new geography but the actual shifting effects, while present (e.g., ``Public transport'' on Fig.~\ref{fig:examples}a and~\ref{fig:examples}d or ``Fish'' on Fig.~\ref{fig:examples}c and~\ref{fig:examples}f), are hard to find;
    \item but different sets have widely varying distributions of labels; this is the main cause of accuracy deteriorating from training set to test sets, and this was our main concern in the challenge;
    \item also, the problem sometimes is further complicated by problems unrelated to domain shift: e.g., on Fig.~\ref{fig:examples}h the model has to recognize a very specific class ``Boeing 737 next generation'', Fig.~\ref{fig:examples}a has label ``Vehicle'' but Fig.~\ref{fig:examples}d does not, and so on; this to some extent explains the low absolute values of final evaluation metrics.
\end{itemize}

The main evaluation metric for the competition was defined as the $F_2$ score:
$$F_2 = \frac{5\times \text{precision} \times \text{recall}}{4\times\text{precision} + \text{recall}},$$
which is the weighted F-measure with recall being twice as important as precision.

The task was to construct and train a model that would work well not only on samples from the same data distribution as in the training set but would also readily generalize to other geographical distributions. During the competition, participants were able to submit their predictions for Stage~1 test set and see evaluation scores on that set. At the end of the challenge, participants uploaded a final model, and a new test set (Challenge Stage~2) was released and scored. Thus, the challenge was to create a model that would be able to generate the best predictions for the Stage~2 test set without any changes in the model or further tuning.

Moreover, there were several technical restrictions imposed on the solutions in the Inclusive Images Challenge:
\begin{itemize}
\item pretrained models were not allowed;
\item training on external datasets was not allowed;
\item predictions should be made based only on the image and not on its metadata (including the geographical location where the picture was taken);
\item the model should be locked and uploaded by the Stage~1 deadline, with no further changes allowed.
\end{itemize}

In summary, the challenge was organized to make the participants train models that would be robust and easy to generalize. However, as we will see below, the similarity between Stage~1 and Stage~2 distributions proved to be higher than probably expected, and this played an important role in the outcome of the competition.

\section{Methods}\label{sec:methods}

Our general pipeline consists of three steps:
\begin{enumerate}[(1)]
\item train a convolutional neural network (CNN) on the training set;
\item adapt the last layer and batch normalization statistics to perform well on tuning labels;
\item train an ensemble of different models.
\end{enumerate}

In the remainder of this section, we describe each step in details.

\subsection{CNN Training}

This step is rather straightforward and similar to standard CNN training. We have found that general parameters of the training set in this problem are very similar to those of ImageNet~\cite{deng2009imagenet}, so as the base classifiers we used convolutional neural networks that show state of the art results on ImageNet. During training, we used the Adam optimizer~\cite{kingma2014adam} with initial learning rate $lr=0.001$, reducing it when there was not any improvement in terms of the validation score (score on the validation set). Basically, we used the \textsc{ReduceLROnPlateau} scheduler with parameters $\mathrm{cooldown}=2$, $\mathrm{patience}=2$, $\mathrm{factor}=0.5$.

During this step, we did not apply any data augmentation because the training set has sufficiently many examples. Nevertheless, we used dropout before the last layer of the networks with dropout probability $p=0.3$.

\subsection{Adapting for New Data Distributions}

Having examined the Stage~1 test set, we found that there is no problem with the distribution of different images. Since the training set is huge, there is a lot of diversity in images, and state of the art networks can generalize and capture different classes very well. However, as the challenge itself suggests, simply using networks trained on the OpenImages dataset to predict labels in the Stage~1 or Stage~2 dataset with a different geographic distribution yields very poor results. 

It has turned out that in order to overcome this problem, proper estimation of the distribution of targets in a test dataset was the key to getting a good result in this task. We decided to use the tuning labels to adapt the last layer of a convolutional neural network trained on the original training set to perform well on the Stage~1 test set. During our experiments, we found that even $1000$ labels can suffice to get a substantial increase in the resulting evaluation metric. In practice, we split these $1000$ labels equally into ten random folds and used a standard cross-validation technique for training. Hence, for each model we obtained ten new models fine-tuned on different subsets of the tuning labels. At the inference stage, we only averaged predictions across these models.

To test our changes against the original training set, we split $100$K images from it as a validation set. Again as expected, we saw a significant drop in the validation score after changing the last layer. It again confirms our hypothesis about the high impact of the distribution of targets. Thus, we decided to use both validation and tuning samples while adapting the last layer, taking for each minibatch a sample from the validation data with probability $\alpha$ and a sample from tuning data with probability $1-\alpha$.

Also, to decrease the chance of overfitting and increase the stability of training we added many different augmentations. For this purpose, we used the \emph{Albumentations} library~\cite{buslaev2018albumentations}, a very fast and efficient implementation of various image augmentation techniques. Table~\ref{tbl:augs} summarizes all augmentations we applied at this stage.

\begin{table}[t]\centering\small
\setlength{\tabcolsep}{5pt}
\begin{tabular}{|l|l|c|}\hline
{\bf Name} & {\bf Description} & {\bf Probability} \\\hline
\textsc{RandomRotate90} & Random rotation by 90 degrees & $0.5$ \\
\textsc{Flip} & Horizontally, vertically or both flips & $0.5$ \\
\textsc{Transpose} & Swapping rows and columns. & $0.5$ \\
\textsc{GaussNoise} & Gaussian noise & $0.1$ \\
\textsc{MedianBlur} & Blurring using a median filter with kernel size 3 & $0.2$ \\
\textsc{RandomShift} & Shifting for a maximum $10\%$ of pixels & $0.5$ \\
\textsc{RandomRotate} & Rotation by a random angle from $0$ to $45$ degrees & $0.5$ \\
\textsc{RandomScale} & Scaling by a random factor from $0.8$ to $1.2$ & $0.5$ \\
\textsc{RandomBrightness} & Brightness changing & $0.15$ \\
\textsc{HueSaturationValue} & Random changing hue, saturation and value & $0.5$ \\\hline
\end{tabular}
\caption{Types of augmentations used while tuning the last layer}\label{tbl:augs}
\end{table}

\subsection{Ensembles}

It is broadly known that uniting different models into an ensemble can often provide a boost in the overall quality of the model. Usually, one of the key factors to winning a data science competition is to find a way of building a strong ensemble. Our solution consists of several groups of models. To reduce the number of hyperparameters, within each group we averaged all models with equal weights. Therefore, the task of constructing the ensemble has been reduced to tuning the weights for averaging different groups, so the number of hyperparameters is equal to the number of such groups.

In practice, we used our validation data and Stage~1 leaderboard to choose the correct weights for the ensemble. We hypothesized that the distribution of targets in Stage~2 will be much more similar to the distribution of Stage~1 than to the training set, but there will still remain a probability that some regions will be identical with those from the training set. Therefore, to construct a more stable solution we used the following procedure:
\begin{itemize}
    \item find sets of weights for which the final ensemble still yields the first place on Stage~1 test data (estimated by the leaderboard scores achieved by our submissions during Stage~1);
\item across all such solutions, find the best one in terms of local validation score.
\end{itemize}

Our hypothesis was based on Figure~\ref{fig:distribution} that was publicly available during the challenge. It was later confirmed by our model winning the Stage~2 competition and by the experiments that we describe below.

\section{Experiments}\label{sec:experiments}

\begin{figure}[t]
    \begin{center}
        \includegraphics[width=\linewidth]{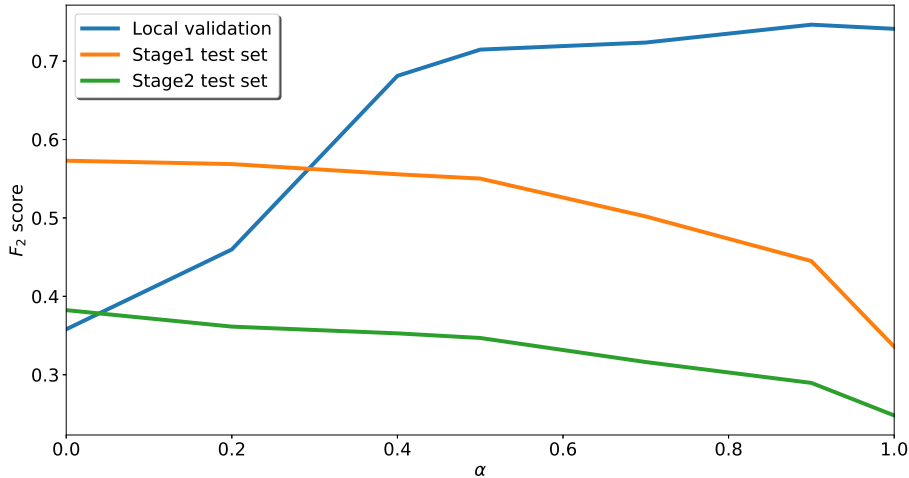}
    \end{center}
    
   \caption{Model scores on the validation set, Stage~1 and Stage~2 test sets as a function of the mixing coefficient $\alpha$ used for the tuning of the last layer.}
\label{fig:dependency}
\end{figure}

During our experiments, all images were downsized to $224\times 224$ pixels as it is a standard resolution for CNN models trained on ImageNet. Also, we have set aside $100$K images from the training set as the local validation set.

Since, as we have already mentioned, the dataset and the task in the Inclusive Images Challenge were very similar to ImageNet classification, in order to speed up experiments and decrease the search space of models we decided to use CNN models that work well on ImageNet:
\begin{itemize}
    \item deep residual networks ResNet101 and ResNet152~\cite{He2016DeepRL};
    \item densely connected convolutional network DenseNet121~\cite{8099726};
    \item the result of neural architecture search with a recurrent network producing the CNN architecture and trained with reinforcement learning, NASNet-A-Large~\cite{Zoph2018LearningTA};
    \item the result of progressive neural architecture search based on sequential model-based optimization, PNASNet-5-Large~\cite{Liu2018ProgressiveNA};
    \item squeeze-and-excitation network SE-Net 154~\cite{Hu2018SqueezeandExcitationN}.
\end{itemize}
Table~\ref{tbl:models} shows the scores for all single models in our experiments. As expected, the validation scores are strongly correlated with the results these models achieve on ImageNet classification.

During our experiments, we found that larger batch size speeds up the convergence of the models; this effect has been explained in~\cite{smith2017don}. Therefore, we used batch size as large as could fit into GPU memory. Using a server with $8\times$NVIDIA Tesla P40, 256 GB RAM and 28 CPU cores, our training process took approximately $30$ days, and inference for Stage~2 data ran in $24$ hours.

\begin{table}[t]\centering\small
\begin{tabular}{|l|c|c|c|c|}\hline
{\bf Network} & {\bf Validation} & {\bf Stage1} & {\bf Stage2} & {\bf ImageNet} \\\hline
DenseNet121~\cite{8099726} & \textbf{0.7159} & 0.3199 & 0.2271 & 0.7498 \\
DenseNet121 (tuned) & 0.3132 & 0.5268 & 0.3368 & - \\\hline
NASNet-A-Large~\cite{Zoph2018LearningTA} & 0.6795 & 0.3145 & 0.2298 & 0.827 \\
NASNet-A-Large (tuned) & 0.3446 & \textbf{0.5443} & 0.3554 & - \\\hline
PNASNet-5-Large~\cite{Liu2018ProgressiveNA} & 0.7129 & 0.3301 & 0.2421 & \textbf{0.829} \\
PNASNet-5-Large (tuned) & 0.3957 & 0.5340 & \textbf{0.3607} & - \\\hline
ResNet101~\cite{He2016DeepRL} & 0.6959 & 0.3189 & 0.2256 & 0.8013 \\
ResNet101 (tuned) & 0.2932 & 0.5209 & 0.3182 & - \\\hline
ResNet152~\cite{He2016DeepRL} & 0.7117 & 0.3201 & 0.2240 & 0.8062 \\
ResNet152 (tuned) & 0.2897 & 0.5239 & 0.3091 & - \\\hline
SE-Net 154~\cite{Hu2018SqueezeandExcitationN} & 0.7151 & 0.3272 & 0.2389 & 0.8132 \\
SE-Net 154 (tuned) & 0.3938 & 0.5226 & 0.3401 & - \\\hline
\end{tabular}
\caption{Single model scores on the validation set, Stage~1 and Stage~2 test sets, and their ImageNet scores (taken from the corresponding references).}\label{tbl:models}
\end{table}

Figure~\ref{fig:dependency} illustrates the scores on validation and test data with different proportions $\alpha$ of the validation data used to tune the last layer of the networks. It clearly shows that choosing lower values of $\alpha$ gives better scores on the test data and a lower score on the validation, which again confirms that target distributions in Stage~1 and Stage~2 are very similar, and the tuning labels are very useful for the Stage~2 part.

In the end, we had five groups of models where each group consisted of models trained with the same $\alpha$. Thus, for the final solution we used a weighted average of these five groups. The weights were chosen based on Stage~1 and validation scores. Table~\ref{tbl:ensembles} shows the final scores for each group and their final weights.

\begin{table}[t]\centering\small
\begin{tabular}{|c|c|c|c|c|}\hline
{\bf Group} & {\bf Validation} & {\bf Stage1} & {\bf Stage2} & {\bf Weight} \\\hline
No tuned & 0.7412 & 0.3358 & 0.2481 & 0.05 \\
Tuned, $\alpha=0$ & 0.3580 & \textbf{0.5730} & \textbf{0.3824} & 0.6 \\
Tuned, $\alpha=0.5$ & 0.7147 & 0.5502 & 0.3469 & 0.3 \\
Tuned, $\alpha=0.9$ & \textbf{0.7465} & 0.4450 & 0.2896 & 0.05 \\\hline
Final ensemble & 0.6253 & \textbf{0.5755} & \textbf{0.3915} & - \\\hline
\end{tabular}
\caption{Ensemble scores on the validation set, Stage~1 and Stage~2 test sets.}\label{tbl:ensembles}
\end{table}

The final ensemble achieved a Stage~2 score of $0.3915$, which was the top scoring entry in the Inclusive Images Challenge at NeurIPS 2018. We also note specifically that the ensembling, while it has allowed us to win the competition, was not the key element to having a good model: our best single model, PNASNet-5-Large, achieved a score corresponding to the $4^{\text{th}}$ place in the challenge.

\section{Conclusion}\label{sec:concl}

In this work, we have presented the winning solution for the Inclusive Images NeurIPS 2018 Challenge~\cite{challenge}. The key components of our solution are the fine-tuning the last layers of base CNN models on a combination of local validation and Stage~1 test sets and an ensemble that includes models trained for several different values of the combination weight $\alpha$ with tuned weights.

In the challenge, this relatively simple approach has proven to be more efficient than attempts based on state of the art domain adaptation methods. This does not mean that domain adaptation is useless, and this success can be mostly attributed to the (probably unintentional) fact that the Stage~2 geographical distribution was quite similar to Stage~1. But the main positive conclusion we can draw from our winning solution is that even a very small labeled set from the target domain, with proper augmentations and ensembling to avoid overfitting, can be extremely useful for transferring pretrained models from one data distribution to another. While domain shift remains a formidable problem, we believe that our approach opens up interesting possibilities for solving this problem in practice, where small labeled datasets from the target domain are usually either available or can be obtained and labeled relatively cheaply.

\bibliographystyle{plain}
\bibliography{references}

\end{document}